\documentclass[11pt,a4paper]{article}
\usepackage[hyperref]{acl2019}
\usepackage[utf8]{inputenc}
\usepackage{graphicx}
\usepackage{gb4ea}
\usepackage{amssymb}
\usepackage{times}
\usepackage{latexsym}
\usepackage{amsmath}
\usepackage{arydshln}

\newcommand{\greenb}{green!40!black}

\usepackage{pifont}%
\newcommand{\cmark}{\textcolor{\greenb}{\ding{51}}}%
\newcommand{\xmark}{\textcolor{red}{\ding{55}}}%

\newcommand{\SR}[1]{\ensuremath{\mathsf{#1}}}
\newcommand{\IntFun}[2]{\ensuremath{I(\SR{#1})=\setof{#2}}}

\newcommand{\setof}[1]{\ensuremath{\left\{#1\right\}}}

\newcommand{\average}[1]{\ensuremath{\langle#1\rangle} }

\newcommand{\revise}[1]{\textcolor{black}{#1}}

\def\aclpaperid{186} 
\aclfinaltrue

\title{Multimodal Logical Inference System for Visual-Textual Entailment}

\author{Riko Suzuki$^1$\\
  {\small \tt suzuki.riko@is.ocha.ac.jp}\And
  Hitomi Yanaka$^{1,2}$\\
  {\small \tt hitomi.yanaka@riken.jp}\AND
  Masashi Yoshikawa$^3$\\
  {\small \tt yoshikawa.masashi.}\\{\small \tt yh8@is.naist.jp}\And
  Koji Mineshima$^1$\\
  {\small \tt mineshima.koji@ocha.ac.jp}\And
  Daisuke Bekki$^1$\\
  {\small \tt bekki@is.ocha.ac.jp}\AND
  $^1$\mbox{\rm Ochanomizu University, Tokyo, Japan}\\
  $^2$\mbox{\rm RIKEN Center for Advanced Intelligence Project, Tokyo, Japan}\\
  $^3$\mbox{\rm Nara Institute of Science and Technology, Nara, Japan}
}

\date{}

\begin{document}
\maketitle

\begin{abstract}
    A large amount of research about multimodal inference across text and vision has been recently developed to obtain visually grounded word and sentence representations. In this paper, we use logic-based representations as unified meaning representations for texts and images and present an unsupervised multimodal logical inference system that can effectively prove entailment relations between them. We show that by combining semantic parsing and theorem proving, the system can handle semantically complex sentences for visual-textual inference.
\end{abstract}

\section{Introduction}
\label{sec:intro}
Multimodal inference across image data and text has the potential to improve understanding information of different modalities and acquiring new knowledge.
Recent studies of multimodal inference provide challenging tasks such as visual question answering~\cite{VQA,Hudson_2019_CVPR,acharya2019tallyqa} and visual reasoning~\cite{suhr-etal-2017-corpus,vu-etal-2018-grounded,xie2018visual}.

Grounded representations from image-text pairs are useful to solve such inference tasks.
With the development of large-scale corpora such as Visual Genome~\cite{krishnavisualgenome} and methods of automatic graph generation from an image~\cite{Xu_2017_CVPR,Qi_2019_CVPR}, we can obtain structured representations for images and sentences such as scene graph~\cite{Johnson2015ImageRU}, a visually-grounded graph over object instances in an image.

While graph representations provide more interpretable representations for text and image than
embedding them into high-dimensional vector spaces~\revise{\cite{NIPS2013_5204,42371}},
there remain two challenges: (i) to capture complex logical meanings such as negation and quantification, and (ii) to perform logical inferences on them.

\begin{figure}[tb]
\centering
\includegraphics[scale=0.27]{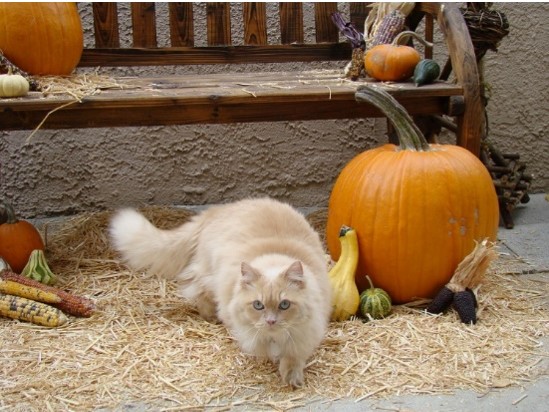}
\raisebox{2.5em}{
\begin{minipage}{11em}\small
\xmark \ \textbf{No} cat is next to a pumpkin.\hfill{(1)}

\smallskip

\xmark \ There are \textbf{at least two} cats.\hfill{(2)}

\smallskip

\cmark \textbf{All} pumpkins are orange.\hfill{(3)}
\end{minipage}
}
\caption{
An example of visual-textual entailment.
An image paired with logically complex statements, namely, negation (1), numeral (2), and quantification (3), leads to a true (\cmark) or false (\xmark) judgement.
}
\label{fig:image-caption}
\end{figure}

For example, 
consider the task of checking
if each statement in Figure
\ref{fig:image-caption}
is true or false under the situation
described in the image.
\revise{The statements (1) and (2) are false,
while (3) is true.}
\revise{To perform this task, it is necessary to handle semantically complex phenomena such as negation, numeral, and quantification.}

To enable such advanced 
visual-textual inferences, it is desirable to build a framework for representing richer semantic contents of texts and images and handling inference between them.
We use logic-based representations as unified meaning representations for texts and images and present an unsupervised inference system
that can 
prove entailment relations between them.
Our visual-textual inference system 
combines semantic parsing via Combinatory Categorial Grammar (CCG; \citet{Steedman2000}) and first-order theorem proving~\cite{JohanModel}.
To describe information in images as logical formulas, we propose a method of transforming graph representations 
into logical formulas,
using the idea of predicate circumscription~\cite{mccarthy1986applications},
which complements information
implicit in images 
\revise{using} the closed
world assumption.
%
Experiments show that our system can perform visual-textual inference with
semantically complex sentences.
\begin{figure*}[!ht]
\begin{center}
\includegraphics[width=\linewidth]{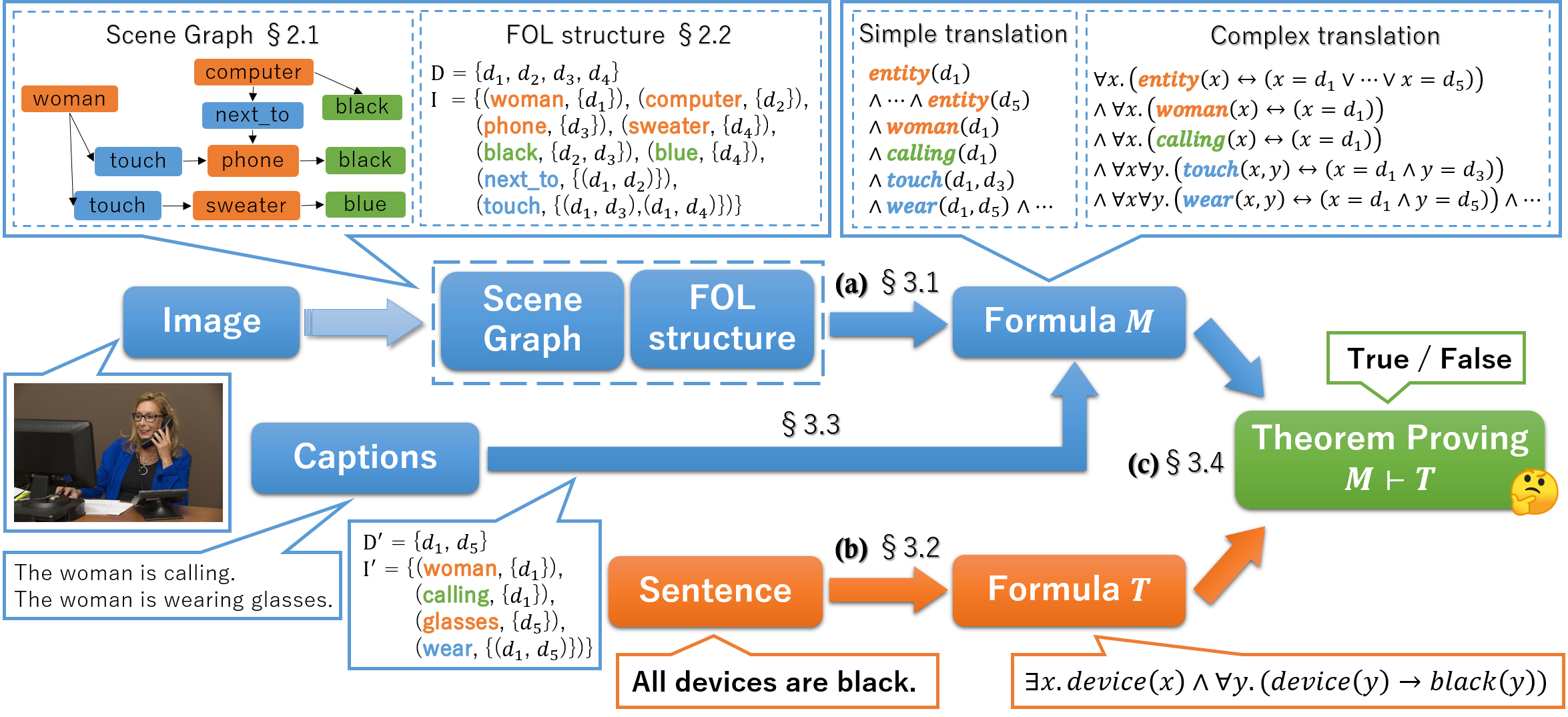}
\caption{Overview of the proposed system.
In this work, we assume the input image is processed into an FOL structure or scene graph a priori.
The system consists of three parts:
(a) \textbf{Graph Translator} converts an image annotated with a scene graph/FOL structure to formula $M$;
(b) \textbf{Semantic parser} maps a sentence to formula $T$ via CCG parsing;
(c) \textbf{Inference Engine} checks whether $M$ entails $T$ by FOL theorem proving.}
\label{fig:overview}
\end{center}
\end{figure*}

\section{Background}
\label{sec:rel}
There are two types of grounded meaning
representations for images:
scene graphs and \revise{first-order logic (FOL)} structures.
Both characterize objects and their
semantic relationships in images.


\subsection{Scene Graph}

A \textit{scene graph}, as proposed in \citet{Johnson2015ImageRU}, is a graphical representation that depicts objects, their attributes, and relations among them occurring in an image.
An example is given
in Figure \ref{fig:overview}.
Nodes in a scene graph
correspond to objects with their categories (e.g.\,\textit{woman}) and
edges correspond to the relationships between
objects (e.g.\,\textit{touch}).
%
%
Such a graphical representation has been shown 
to be useful in high-level tasks such as image retrieval~\cite{Johnson2015ImageRU, W15-2812}
and visual question answering~\cite{TeneyLH17}.
Our proposed method builds on 
the idea that these graph
representations can be translated
into logical formulas and be used in 
complex logical reasoning.


\subsection{FOL Structure}
\label{subsec:fol_structure}

In logic-based approaches
to semantic representations,
\textit{FOL structures} (also called \textit{FOL models}) are used to represent
semantic information in images~\citep{hurlimann-bos:2016:VL16},
An FOL structure is a pair $(D, I)$
where $D$ is a domain (also called \textit{universe}) consisting of
all the entities in an image and 
$I$ is an interpretation function
that maps a 1-place predicate
to a set of entities
and a 2-place predicate to a set of
pairs of entities, and so on;
for instance,
we write 
$\IntFun{man}{d_1}$ if the entity $d_1$ is a man, and
$\IntFun{next\_to}{(d_1,d_2)}$ 
if $d_1$ is next to $d_2$.
%
FOL structures have
clear correspondence with the graph
representations of images
in that they both 
capture the categories, attributes and relations holding of the entities
in an image.
For instance,
the FOL structure and scene graph in the upper left of Figure \ref{fig:overview} have exactly the same information.
Thus, the translation from graphs
to formulas can also work for FOL structures (see \S\ref{subsec:graph_formula}).

\section{Multimodal Logical Inference System}
\label{sec:method}


Figure~\ref{fig:overview} shows the overall picture of the proposed system.
We use formulas of
FOL with equality
as unified semantic representations
for text and image information.
We use 1-place and 2-place predicates
for representing attributes and relations,
respectively.
The language of FOL consists of
(i) a set of atomic formulas,
(ii) equations of the form $t\!=\!u$,
and (iii) complex formulas composed
of negation ($\neg$),
conjunction ($\wedge$),
disjunction ($\vee$),
implication ($\to$), and
universal and existential quantification ($\forall$ and $\exists$).
The expressive power of the FOL language
provides a structured representation
that captures not only objects and their
semantic relationships
but also those complex
expressions including
negation, quantification
and numerals.

The system takes as input
an image $I$ and a sentence $S$
and determines whether $I$ entails $S$,
\revise{in other words,} $S$ is true with respect
to the situation described in $I$.
In this work, we assume the input image $I$ is processed into a scene graph/FOL structure $G_I$ using an off-the-shelf converter~\cite{Xu_2017_CVPR,Qi_2019_CVPR}.

To determine entailment relations between
sentences and images, we proceed in three
steps.
First, \textbf{graph translator}
maps \revise{a graph} $G_I$
to a formula $M$.
We develop two ways of translating
graphs to FOL formulas
(\S\ref{subsec:graph_formula}).
Second, \textbf{semantic parser}
takes a sentence $S$ as input
and return a formula $T$ via
CCG parsing.
We improve a semantic parser in CCG for
handling numerals and quantification
(\S\ref{subsec:sentence_formula}).
Additionally, 
we develop a method 
for utilizing image captions
to extend $G_I$ with information obtainable from their logical formulas (\S\ref{subsec:cap_fol}).
Third, \textbf{inference engine}
checks whether $M$ entails $T$, written
$M \vdash T$,
using FOL theorem prover (\S\ref{subsec:inference}).
Note that FOL theorem provers can accept
multiple premises, $M_1, \ldots, M_n$,
converted from images and/or sentences
and check if $M_1, \ldots, M_n \vdash T$
holds or not. Here we focus on
single-premise visual inference.

\subsection{Graph Translator}\label{subsec:graph_formula}

We present two ways of translating
graphs (or equivalently, FOL structures) to formulas:
a simple translation (Tr$_s$)
and a complex translation (Tr$_c$).
These translations are defined in Table \ref{table:translation}.
%
\begin{table}
{\small
\setlength{\leftskip}{-1em}
\begin{description}
\item Tr$_s(D)$ = $\SR{entity}(d_1)\land \ldots \land \SR{entity}(d_n)$
\vspace{-0.3em}
\item Tr$_s(P)$ = $P(d_1)\land \ldots \land P(d_{n'})$
\vspace{-0.3em}
\item Tr$_s(R)$ = $R(d_{i_1}, d_{j_1})\land \ldots \land R(d_{i_n}, d_{j_n})$
\vspace{-0.3em}
\item Tr$_c(D)$ = $\forall x. (\SR{entity}(x) \leftrightarrow (x = d_1 \lor \ldots \lor x = d_n))$
\vspace{-0.3em}
\item Tr$_c(P)$ = $\forall x. (P(x) \leftrightarrow (x = d_1 \lor \ldots \lor x = d_{n'}))$
\vspace{-0.3em}
\item Tr$_c(R)$ = $\forall x \forall y. (R(x,y) \leftrightarrow ((x = d_{i_1} \land y = d_{j_1}) \lor \ldots \\
\hspace{2em} \lor (x = d_{i_m} \land y = d_{j_m})))$
\end{description}}
\vspace{-0.6em}
\caption{Definition of two types of translation, TR$_s$ and TR$_c$.
Here we assume that $D = \setof{d_1, \ldots, d_n}$,
$P = \setof{d_1, \ldots, d_{n'}}$, and
$R = \setof{(d_{i_1}, d_{j_1}), \ldots,
(d_{i_m}, d_{j_m})}$.}
\label{table:translation}
\end{table}
For example,
consider a graph
consisting of the domain $D = \setof{d_1, d_2}$,
where we have $\SR{man}(d_1), \SR{hat}(d_2), \SR{red}(d_2)$
as properties
and $\SR{wear}(d_1, d_2)$ as relations.
The simple translation TR$_s$
gives the formula (S) below,
which simply conjoins all the atomic information.
%
%
\begin{exe}
\setlength{\leftskip}{-0.5em}
\exi{(S)} $\SR{man}(d_1) \land \SR{hat}(d_2) \land \SR{red}(d_2) \land \SR{wear}(d_1,d_2)$
\end{exe}
However, this does not capture
the \textit{negative} information
that $d_1$ is the only entity
that has the property \SR{man};
similarly for the other predicates.
To capture it,
we use the complex translation Tr$_c$,
which gives the following formula:

\begin{exe}
\exi{(C)} $\forall x. (\SR{man}(x) \leftrightarrow x = d_1) \land \\
                        \forall y. (\SR{hat}(y)\leftrightarrow y = d_2) \land \\
                        \forall z.(\SR{red}(z) \leftrightarrow z = d_2) \land \\
                        \forall x \forall y. (\SR{wear}(x,y) \leftrightarrow (x = d_1\land y = d_2))$
\end{exe}



\noindent
This formula says that $d_1$ is the only
man in the domain,
$d_2$ is the only hat in the domain, and
so on.
This way of translation
can be regarded as an instance of Predicate Circumscription~\cite{mccarthy1986applications},
which complement negative information
\revise{using}
the closed world assumption.
The translation  Tr$_c$
is useful for handling formulas with negation and universal quantification.

One drawback here is that
since (C) involves complex formulas, it 
\revise{increases}
the computational cost in theorem proving.
To remedy this problem,
we use two types of translation selectively,
depending on the polarity of the formula to
be proved.
%
Table \ref{fig:polarity}
shows the definition to classify each FOL formula $A \in \mathcal{L}$ into
positive $(\mathcal{P})$
and negative $(\mathcal{N})$ one.
For instance,
the formulas $\exists x \exists y. (\SR{cat}(x) \wedge \SR{dog} \wedge \SR{touch}(x,y))$,
which correspond to 
\textit{A cat touches a dog},
is a positive formula,
while
$\neg \exists x. (\SR{cat}(x) \wedge \SR{white}(x))$,
which corresponds to 
\textit{No cats are white},
is a negative formula.


\begin{table}
{\small
\begin{enumerate}
\setlength{\itemsep}{-0.2em}
\item $A \in \mathcal{P},\, \neg A \in \mathcal{N}$, if $A$ is an atomic formula. 
\item $A,\, \neg A \in \mathcal{P}$, if $A$ is an equation of the form $t\!=\!u$.
\item $A \wedge B,\, A \vee B \in \mathcal{P}$, if $A \in \mathcal{P}$ and $B \in \mathcal{P}$.
\item $A \wedge B,\, A \vee B \in \mathcal{N}$, if $A \in \mathcal{N}$ or $B \in \mathcal{N}$.
\item $A \to B \in \mathcal{P}$, if $A \in \mathcal{N}$ and $B \in \mathcal{P}$.
\item $A \to B \in \mathcal{N}$, if $A \in \mathcal{P}$ or $B \in \mathcal{N}$.
\item $\forall x. A,\, \exists x. A \in \mathcal{P}$, if $A \in \mathcal{P}$.
\item $\forall x. A,\, \exists x. A \in \mathcal{N}$, if $A \in \mathcal{N}$.
\end{enumerate}}
\vspace{-0.6em}
\caption{Positive $(\mathcal{P})$ and negative $(\mathcal{N})$ formulas}
\label{fig:polarity}
\end{table}

\subsection{Semantic Parser}
\label{subsec:sentence_formula}

We use ccg2lambda~\cite{mineshima-EtAl:2015:EMNLP},
a semantic parsing system based on CCG to convert sentences to formulas,
and extend it to handle numerals and quantificational
sentences.
In our system, a sentence with numerals,
e.g., 
\textit{There are (at least) two cats},
is compositionally mapped to the following FOL formula:
%
\begin{exe}
\exi{(\textsf{Num})} $\exists x \exists y. (\SR{cat}(x) \land \SR{cat}(y) \land (x \neq y))$
\end{exe}
Also, to capture the existential import of
universal sentences,
the system maps
the sentence \textit{All cats are white}
to the following one:
%
\begin{exe}
\exi{(\textsf{Q})} \revise{$\exists x. \SR{cat}(x) \land \forall y.(\SR{cat}(y) \rightarrow \SR{white}(y))$}
\label{fol:allcats2}
\end{exe}

\subsection{Extending Graphs with Captions}
\label{subsec:cap_fol}

Compared with images,
captions can describe a variety of properties and relations
other than spatial and visual ones.
By integrating caption information into FOL structures, we can obtain semantic representations reflecting
relations that can be described only in the caption.

We convert captions into 
\revise{FOL structures (= graphs)}
using our semantic parser.
We only consider the cases where
the formulas obtained are
composed of existential quantifiers and conjunctions. 
For extending \revise{FOL structures}
with caption information, it is necessary to analyze co-reference
between the entities occurring in 
sentences and images.
We add a new predicate to 
an \revise{FOL structure}
if the co-reference is uniquely determined.

As an illustration, consider the captions 
and the FOL structure $(D, I)$ which represents the image shown in Figure \ref{fig:overview}.\footnote{
Note that there is a unique correspondence between FOL structures and scene graphs. For the sake of illustration, we use FOL structures in this subsection.
}
The captions, (\ref{ex:call}a) and (\ref{ex:glass}a), are mapped to the formulas (\ref{ex:call}a) and (\ref{ex:glass}b), respectively, via semantic parsing.

\begin{exe}
\ex \label{ex:call}
\begin{xlist}
\ex The woman is calling.
\ex $\exists x. (\SR{woman}(x) \land \SR{calling}(x))$
\end{xlist}
\ex \label{ex:glass}
\begin{xlist}
\ex The woman is wearing glasses.
\ex $\exists x \exists y. (\SR{woman}(x)\land\SR{glasses}(y)$\\
\quad $\land\ \SR{wear}(x,y))$
\end{xlist}
\end{exe}


\noindent
Then, the information in (\ref{ex:call}b) and (\ref{ex:glass}b) can be added to $(D, I)$, because there is only one woman $d_1$ in $(D, I)$ and thus the co-reference between \textit{the woman} in the caption and the entity $d_1$ is uniquely determined.
Also, a new entity $d_5$ for glasses
is added because there are no such entities in the structure $(D, I)$.
Thus we obtain the following new structure ($D^*, I^*$) extended with the information in the captions.
\begin{align*}
D^* := &\ D \cup \{d_5\} \\
I^* := &\ I \cup \{(\SR{glasses},\{d_5\}),(\SR{calling},\{d_1\}), \\
& (\SR{wear},\{(d_1, d_5)\})\}
\end{align*}
%
%


\subsection{Inference Engine}
\label{subsec:inference}

Theorem prover is a method for judging whether
a formula $M$ entails a formula $T$.
We use Prover9\footnote{http://www.cs.unm.edu/~mccune/prover9/}
as an FOL prover for inference.
We set timeout (10 sec)
to judge that $M$ does not entail $T$.



\section{Experiment}
\label{sec:experiment}

We evaluate the performance of the proposed visual-textual inference system.
Concretely, we formulate our task as image retrieval using query 
\revise{sentences} and evaluate the performance in terms of the number of correctly returned images.
In particular, we focus on semantically complex sentences containing numerals, quantifiers, and negation, which are difficult for existing graph representations to handle.

\paragraph{Dataset:}\
We use two datasets: Visual Genome~\cite{krishnavisualgenome}, which contains pairs of scene graphs and images, and GRIM dataset~\cite{hurlimann-bos:2016:VL16}, which annotates an FOL structure of an image and two types of captions (true and false sentences 
with respect to the image).
Note that our system is fully unsupervised and does not require any training data; \revise{in the following,} we describe only test set creation procedure.

\begin{table}[t]
\centering
\scalebox{0.85}{
\begin{tabular}{l|l} \hline
    \multicolumn{1}{c|}{Pattern} &  \multicolumn{1}{|c}{Phenomena}\\ \hline
    There is a \average{attr}\average{attr}\average{obj}. & Con \\
    There are at least \average{number}\average{obj}. & Num \\
    All \average{obj} are \average{attr}. & Q \\
    \average{obj}\average{rel}\average{obj}. & Rel \\
    No \average{obj} is \average{attr}. & Neg \\
    All \average{obj}\average{attr} or \average{attr}. & Con, Q \\
    Every \average{obj} is not \average{rel}\average{obj}. & Num, Rel, Neg \\ \hline
\end{tabular}
}
\caption{Examples of sentence templates. \average{obj}: objects, \average{attr}: attributes, \average{rel}: relations.}
\label{table:pattern}
\end{table}

\begin{table}[t]
\begin{center}
\scalebox{0.75}{
\begin{tabular}{l|l|r} \hline
    \multicolumn{1}{c|}{Sentences} & \multicolumn{1}{|c|}{Phenomena} & \multicolumn{1}{|c}{Count} \\ \hline
    There is a long red bus. & Con & 3 \\
    There are at least three men. & Num & 32 \\
    All windows are closed. & Q & 53 \\
    Every green tree is tall. & Q & 18 \\
    A man is wearing a hat. & Rel & 12 \\
    No umbrella is colorful. & Neg & 197 \\
    There is a train which is not red. & Neg & 6 \\
    There are two cups or three cups. & Con, Num & 5 \\
    All hairs are black or brown. & Con, Q & 46 \\
    A gray or black pole has two signs. & Con, Num, Rel & 6 \\
    Three cars are not red. & Num, Neg & 28 \\
    All women wear a hat. & Q, Rel & 2 \\
    A man is not walking on a street. & Rel, Neg & 76 \\
    A clock on a tower is not black. & Rel, Neg & 7 \\
    Two women aren't having black hair. & Num, Rel, Neg & 10 \\
    Every man isn't eating anything. & Q, Rel, Neg & 67 \\ \hline
\end{tabular}
}
\caption{
Examples of query sentences In \S\ref{subsec:experiment_sg}; Count shows the number of images describing situations under which each sentence is true.
}
\label{table:sentences}
\end{center}
\end{table}

For the experiment using Visual Genome, we randomly extracted 200 images as test data, and a separate set of 4,000 scene graphs for creating query sentences;
we made queries by the following steps.
First, we prepared sentence templates focusing on five types of linguistic phenomena: logical connective (\textbf{Con}), numeral (\textbf{Num}), quantifier (\textbf{Q}), relation (\textbf{Rel}) and negation (\textbf{Neg}).
See Table~\ref{table:pattern} for the templates.
Then, we manually extracted object, attribute and relation types from the frequent ones (appearing more than 30 times) in the extracted 4,000 graphs, and created queries by replacing \average{obj}, \average{attr} and \average{rel} in the templates with them.
As a result, we obtained 37 semantically complex queries as shown in Table~\ref{table:sentences}.
To assign correct images to each query,
two annotators judged whether each of the test images entails the query sentence.
If the two judgments disagreed, the first author decided the correct label.

In the experiment using GRIM, we adopted the same procedure to create a test \revise{dataset} and obtained 19 query sentences and 194 images.

One of the issues in this dataset is that annotated FOL structures contain only spatial relations such as \textit{next\_to} and \textit{near};
to handle queries containing general relations such as \textit{play} and \textit{sing}, our system needs to utilize annotated captions (\S\ref{subsec:cap_fol}). To evaluate if our system can effectively extract information from captions,
we split \textbf{Rel} of above linguistic phenomena into spatial relation (\textbf{Spa-Rel}; relations about spatial information) and general relation (\textbf{Gen-Rel}; other relations), and report the scores separately in terms of these categories.


\subsection{Experimental Results on Visual Genome} \label{subsec:experiment_sg}
Firstly, we evaluate the performance in terms of our {\bf Graph translator}'s conversion algorithm.
As described in \S\ref{subsec:graph_formula}, there are two translation algorithms; simple one that conjunctively enumerates all relation in a graph (\textbf{\textsc{Simple}} in the following), and one that selectively employs translation based on Predicate Circumscription (\textbf{\textsc{Hybrid}}).

\begin{table}[t]
\centering
\begin{tabular}{cccc} \hline
    \multicolumn{2}{c}{Phenomena (\#)} &
    \textsc{Simple} & 
    \textsc{Hybrid} \\ \hline
    Con & (17) & 36.40 & 41.66 \\
    Num & (9) & 43.07 & 45.45\\
    Q & (9) & 8.59 & \textbf{28.18}\\
    Rel & (11) & 25.13 & \textbf{35.10}\\
    Neg & (11) & 66.38 & \textbf{73.39}\\ \hline
\end{tabular}
\caption{Experimental results on Visual Genome (F1). ``\#'' stands for the number of query sentences categorized into that phenomenon.}
\label{table:result}
\end{table}

\begin{figure}[!ht]
\begin{center}
\scalebox{0.95}{
\begin{tabular}{c}
    \begin{minipage}{\hsize}
    \begin{center}
        \includegraphics[width=\linewidth]{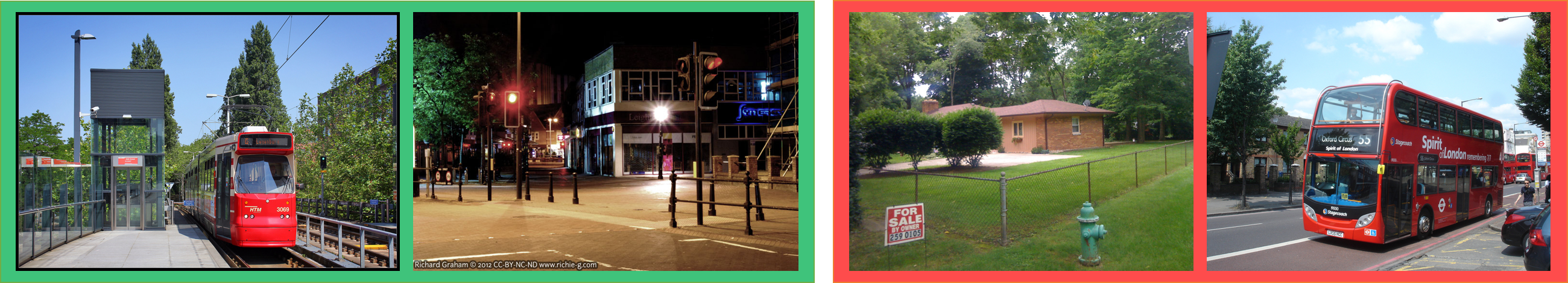}
        \hspace{2cm} (a) \textit{Every green tree is tall.}
    \end{center}
    \end{minipage}\\\\
    \begin{minipage}{\hsize}
        \begin{center}
        \includegraphics[width=\linewidth]{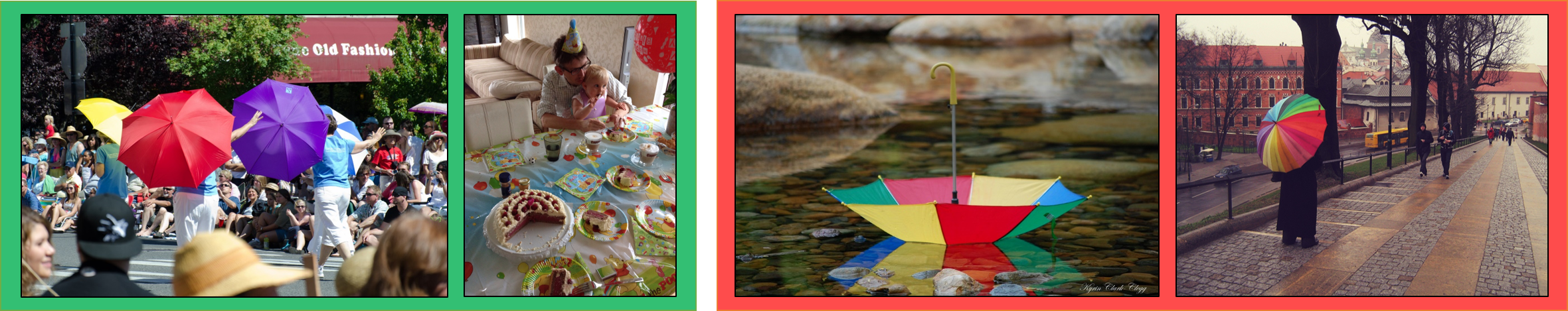}
        \hspace{2cm} (b) \textit{No umbrella is colorful.}
    \end{center}
    \end{minipage}\\
\end{tabular}
}
\caption{Predicted images of our system; Images in green entail the queries, while those in red do not.}
\label{fig:example}
\end{center}
\end{figure}

Table~\ref{table:result} shows image retrieval scores per linguistic phenomenon, macro averages of F1 scores of queries labeled with the respective phenomena.
\textsc{Hybrid} shows better performance for all phenomena than \textsc{Simple} one,
improving by 19.59\% on \textbf{Q}, 9.97\% on \textbf{Rel} and 7.01\% on \textbf{Neg}, over \textsc{Simple},
suggesting that the proposed complex translation is useful for inference using semantically complex sentences including quantifier and negation.
Figure~\ref{fig:example} shows retrieved results for a query (a) {\it Every green tree is tall} and (b) {\it No umbrella is colorful}, each containing universal quantifier and negation, respectively.
Our system successfully performs inference on these queries, returning the correct images, while excluding wrong ones (note that the third picture in (a) contains short trees).

\paragraph{Error Analysis:}\
One of the reasons for the lower F1 of \textbf{Q} is the gap of annotation rule between Visual Genome and our test set.
Quantifiers in natural
language often involve
vagueness~\cite{pezzelle-etal-2018-comparatives}.
for example, the interpretation of \textit{everyone}
depends on what counts as an entity
in the domain. Difficulties in 
fixing the interpretation of quantifiers
caused the lower performance.
%

The low F1 in \textbf{Rel} is primarily due to lexical gaps between formulas of a query and an image.
For example, sentences {\it All women wear a hat} and {\it All women have a hat} are the same in their meaning.
However, if a scene graph contains only \textit{wear} relation, our system can handle the former query, while not the other.
In future work, we will extend our system with a knowledge insertion mechanism~\cite{martinezgomez-EtAl:2017:EACLlong}.

\subsection{Experimental Results on GRIM}
We test our system on GRIM dataset.
As noted above, the main issue on this dataset is the lack of relations other than spatial ones.
We evaluate if our system can be enhanced using \revise{the} information contained in captions.
The F1 scores of the {\textsc Hybrid} system with captions are the same with the one without captions on the sets except for \textbf{Gen-Rel};\footnote{
    \textbf{Con}: 91.41\%,
    \textbf{Num}: 95.24\%,
    \textbf{Q}: 78.84\%,
    \textbf{Spa-Rel}: 88.57\%,
    \textbf{Neg}: 62.57\%.
}
on the subset, the F1 score of the former improves by 60\% compared to the latter,
which suggests that captions can be integrated into FOL structures for the improved performance.


\section{Conclusion}
\label{sec:conclusion}

We have proposed a logic-based system to achieve advanced visual-textual inference,
demonstrating the importance of building a framework for representing \revise{the richer semantic content} of texts and images.
In the experiment, we have shown that our CCG-based pipeline system, consisting of graph translator, semantic parser and inference engine, can perform visual-textual inference with semantically complex sentences, without requiring any supervised data.

\section*{Acknowledgement}
We thank the two anonymous reviewers for their encouragement and insightful comments.
This work was partially supported by JST CREST Grant Number JPMJCR1301, Japan.

\bibliographystyle{acl_natbib}
\small
\bibliography{MyLibrary}

\end{document}